# Assessing the players' performance in the game of bridge: A fuzzy logic approach

**Michael Gr. Voskoglou**

School of Technological Applications, Graduate Technological Educational Institute (T. E. I.) of Western Greece, Patras, Greece
E-mail: mvosk@hol.gr

**Abstract**. Contract bridge occupies nowadays a position of great prestige being, together with chess, the only mind sports officially recognized by the International Olympic Committee. In the present paper an innovative method for assessing the total performance of bridge-players' belonging to groups of special interest (e.g. different bridge clubs during a tournament, men and women, new and old players, etc) is introduced, which is based on principles of fuzzy logic. For this, the cohorts under assessment are represented as fuzzy subsets of a set of linguistic labels characterizing their performance and the centroid defuzzification method is used to convert the fuzzy data collected from the game to a crisp number. This new method of assessment could be used informally as a complement of the official bridge-scoring methods for statistical and other obvious reasons. Two real applications related to simultaneous tournaments with pre-dealt boards, organized by the Hellenic Bridge Federation, are also presented, illustrating the importance of our results in practice.

## 1. Introduction

In this section we shall give a brief description of the fundamentals of the game of contract bridge and we shall discuss the advantages of using principles of fuzzy logic in the assessment procedures in general.

### 1.1 The game of bridge

Bridge is a card game belonging to the family of trick-taking games. It is a development of *Whist*, which had become the dominant such game enjoying a loyal following for centuries.

In 1904 *Auction Bridge* was developed, in which the players bid in a competitive auction to decide the contract and declarer. The object became to make at least as many tricks as were contracted for and penalties were introduced for failing to do so.

The modern game of *Contract Bridge* was the result of innovations to the scoring of auction bridge suggested by *Harold Stirling Vanderbilt* (USA, 1925) and others. Within a few years contract bridge had so supplanted the other forms of the game that "bridge" became synonymous with "contract bridge."

*Rubber Bridge* is the basic form of contract bridge, played by four players. Informal social bridge games are often played this way. *Duplicate Bridge* is the game usually played in clubs, tournaments and matches. The game is basically the same with the rubber bridge, but the luck element is reduced by having the same deals replayed by different sets of players. At least eight players (in two tables) are required for this. There are also some significant differences in the scoring.

Bridge occupies nowadays a position of great prestige being, together with chess, the only *mind sports* (i.e. games or skills where the mental component is more significant than the physical one) officially recognized by the International Olympic Committee. Millions of people play bridge worldwide, not only in clubs, tournaments and championships, but also on line (e.g. [1]) and with friends at home, making it one of the world's most popular card games. The *World Bridge Federation (WBF)* is the international governing body of contract bridge. WBF was formatted in August 1958 by delegates from Europe, North and South America and its membership now comprises 123 National Bridge Organizations, with about 700000 affiliated members.

In the standard 52-card deck used in bridge, the ace is ranked highest followed by the king, queen, and jack (all the above cards called *honours*) and the spot-cards from ten down through to the two. Suit denominations also have a rank order with no trump (NT) being highest followed by spades (SP), hearts (H), diamonds (D) and clubs (CL).

There are four players in each table, in two fixed partnerships. Partners sit facing each other. It is traditional to refer to the players according to their position at the table as North (N), East (E), South (S) and West (W). So N and S are partners against E and W.

An almost essential tool for playing bridge is the *board* containing four pockets, one for each player, marked by N, E, W and S respectively; 13 play cards are placed in each pocket. Each board carries a number to identify it and has marks showing the *dealer* (i.e. the player who starts the bidding) and whether each of the two playing sides is *vulnerable* or not. A side which is vulnerable is subject to higher bonuses and penalties than one that is not.

At the beginning of the game the cards are shuffled, dealt and placed in the pockets of each board. In some competitions boards are pre-dealt prior to the competition, especially if the same hands are to be played at many locations; for example in a large national or international tournament. Mechanical dealing machines or special computer software are usually used for this purpose.

Each session (*hand*) of the game is progressing through the following phases: *Bidding (or auction), play of the cards* and *scoring the results*.

Bidding is based on the premise that the lowest available to bidders starts with the proposition to take seven tricks, i.e. one cannot contract to make less than seven tricks. Given this, the bidding is said to start at the one-level when contracting for a total of seven tricks, at the two-level for eight tricks and so on to the seven-level to contract to take all thirteen tricks. Thus, there are 35 possible contracts, five at each of the seven levels. The dealer begins the bidding, and the turn to speak passes clockwise. At each turn a player may either make a bid or pass. It is also possible to 'double' an opponent's bid, or to 'redouble' the opponent's 'double', thus increasing the score of the bid when won, and the penalties, when lost. If someone then bids higher, any previous 'double' or 'redouble' are cancelled. If all four players pass on their first turn to speak the hand is said to be *passed out*. The cards are thrown in and the next board is played. If anyone bids, then the auction continues until there are three passes in succession, and then stops. In this case the last bid becomes the *contract*.

The team who made the final bid will now try to make the contract. The first player of this team who mentioned the denomination (suit or no trumps) of the contract becomes the *declarer*. The declarer's partner is known as the *dummy*. The player to the left of the declarer leads to the first trick and may play any card. Immediately after this opening lead, the dummy's cards are exposed. Play proceeds clockwise. Each of the other three players in turn must, if possible, play a card of the same suit that the leader played. A player with no card of the suit led may play any card. A trick consists of four cards, one from each player, and is won by the highest trump in it, or if no trumps were played by the highest card of the suit led. The winner of a trick leads to the next, and may lead any card. Dummy takes no active part in the play of the hand. Whenever it is dummy's turn to play, the declarer must say which of dummy's cards is to be played. When dummy wins a trick, the declarer specifies which card dummy should lead to the next trick.

When the play ends, the score is determined by comparing the number of tricks taken by the declaring side to the number required to satisfy the contract.

A match can be played among *teams* (two or more) of four players (two partnerships). At the end of the match in this case the result is the difference in *International Match Points (IMPs)* between the competing teams and then there is a further conversion, in which some fixed number of *Victory Points (VPs)* is appointed between the teams. It is worth to notice that the table converting IMPs to VPs has been obtained through a rigorous mathematical manipulation [4].

However, the game usually played in tournaments is among fixed partnerships or *pairs*. For a pairs event a minimum of three tables (6 pairs, 12 players) is needed, but it works better with more players. Generally you play two or three boards at a table - this is called a *round* - and then one or both pairs move to another table and play other boards against other opponents. The score for each hand is recorded to a *score sheet*, which is kept folded in a special pocket of the board provided for this purpose, so that previous scores could not be read before the board has been played. North is then responsible for entering the result and showing the completed sheet to East-West to check that it has been done correctly. Each pair has an identity number, which must also be entered on the score sheet, to show whose result it is. At the end of the game each score sheet will contain the results of all the pairs who have played that board. The score sheets are then collected by the organisers and the scores are compared. The usual method of scoring in a pairs' competition is in *match points*. Each pair is awarded two match points for each pair who scored worse than them on that board, and one match point for each pair who scored equally. The total number of match points scored by each pair over all the boards is calculated and it is converted to a percentage. The pair succeeding the highest percentage wins the game. However, IMPs are also used as a method of scoring in special cases, in which the difference of each pair's IMPs is usually calculated with respect to the mean number of IMPs of all pairs.

There are also several *conventions* that can be played between the partners. However, a full description of the rules and techniques of bridge is out of the purposes of the present paper.

There are very many books written about bridge, the most famous being probably the book [6] of *Edgar Kaplan* (1925-1997), who was an American bridge player and one of the principal contributors to the game. Kaplan's book was translated in many languages and was reprinted many times since its first edition in 1964; for instance, [7] is one of the recent unabridged republications of it. There is also a fair amount of bridge-related information on the Internet. For the history, the fundamentals and a detailed description of the rules of the game the reader may look at the web sites [2-3], etc.

**1.2 Fuzzy logic as a tool in assessment procedures**

There used to be a tradition in science and engineering of turning to probability theory when one is faced with a problem in which uncertainty plays a significant role. This transition was justified when there were no alternative tools for dealing with the uncertainty. Today this is no longer the case. *Fuzzy logic*, which is based on fuzzy sets theory introduced by Zadeh [19] in 1965, provides a rich and meaningful addition to standard logic. A real test of the effectiveness of an approach to uncertainty is the capability to solve problems which involve different facets of uncertainty. Fuzzy logic has a much higher problem solving capability than standard probability theory. Most importantly, it opens the door to construction of mathematical solutions of computational problems which are stated in a natural language.

The applications which may be generated from or adapted to fuzzy logic are wide-ranging and provide the opportunity for modelling under conditions which are inherently imprecisely defined, despite the concerns of classical logicians (e.g. see Chapter 6 of [8], [11], [12] and its relevant references, [13-15], [17], etc).

The methods of assessing the individuals' performance usually applied in practice are based on principles of the bivalent logic (yes-no). However these methods are not probably the most suitable ones. In fact, fuzzy logic, due to its nature of including multiple values, offers a wider and richer field of resources for this purpose. This gave us several times in the past the impulsion to introduce principles of fuzzy logic in assessing the performance of student groups in learning mathematics and problem solving (e.g. see [10], [12-13], [16-18], etc.). In this paper we shall use fuzzy logic in assessing the total performance of bridge players' belonging to sets of special interest (e.g. different bridge clubs during a tournament, men and women, new and old players, etc).

The rest of the paper is organized as follows: In the next section we develop our new assessment method, which is based on principles of fuzzy logic. In section three we present two real applications illustrating the importance of our method in practice. Finally the last section is devoted to conclusions and discussion on future perspectives of research on this area.

For general facts on fuzzy sets we refer freely to the book [8].

## 2. The assessment method

As we have already seen in the previous section, in a game of duplicate bridge the performance of each element (pair or team) is characterized by using either match points or IMPs. However, apart from the above official scoring methods, it is useful sometimes, for statistical or other reasons, to assess the total performance of certain sets of playing elements (single players, pairs, or teams) appearing to have a special interest. For example, this happens, when one wants to compare the performance of two or more clubs participating in a big tournament, the performance of male and female players or of old and young players, etc.

One way to do this is by calculating the means of the official scores obtained by the elements of the corresponding sets (*mean performance*). Here, we shall use principles of fuzzy logic in developing an alternative method of assessment, according to which the higher is an element's performance the more its "contribution" to the corresponding set's total performance (*weighted performance*).

For this, we consider as set of the discourse the set $U = \{A, B, C, D, F\}$ of linguistic labels characterizing the playing elements' performance, where A characterizes an excellent performance, B a very good, C a good, D a mediocre and F an unsatisfactory performance respectively. Obviously, the above characterizations are fuzzy depending on the user's personal criteria, which however must be compatible to the common logic, in order to model the real situation in a worthy of credit way.

In case of a pairs' competition, for example, with match points as the scoring method and according to the usual standards of duplicate bridge, we can characterize the pairs' (or the players' individually) performance, according to the percentage of success, say p, achieved by them, as follows:

- Excellent (A), if $p > 65\%$.
- Very good (B), if $55\% < p \leq 65\%$.
- Good (C), if $48\% < p \leq 55\%$.
- Mediocre (D), if $40\% \leq p \leq 48\%$.
- Unsatisfactory (F), if $p < 40\%$.

In an analogous way one could characterize the teams' (or pairs') performance with respect to the VPs, gained in bridge games played with IMPs.

Assume now that one wants to assess the total performance of a special set, say S, of n playing pairs (or players'), where n is an integer, $n \geq 2$. We are going to *represent S as a fuzzy subset of U*. For this, if $n_A$, $n_B$, $n_C$, $n_D$ and $n_F$ denote the number of pairs/players of S that had demonstrated an excellent, very good, good, mediocre and unsatisfactory performance respectively at the game, we define the *membership function* $m : U \to [0, 1]$ in terms of the frequencies, i.e. by $m(x) = \frac{n_x}{n}$, for each x in U. Then S can be written as a fuzzy subset of U in the form: $S = \{(x, \frac{n_x}{n}): x \in U\}$.

In converting the fuzzy data collected from the game we shall make use of the defuzzification technique known as the *centroid method*. According to this method, the centre of gravity of the graph of the membership function involved provides an alternative measure of the system's performance. The application of the centroid method in practice is simple and evident and, in contrast to other defuzzification techniques in use, like the measures of uncertainty (for example see [12] and its relevant references, or [15]), needs no complicated calculations in its final step. The techniques that we shall apply here have been also used earlier in [9], [14], [16], etc.

The first step in applying the centroid method is to correspond to each $x \in U$ an interval of values from a prefixed numerical distribution, which actually means that we replace U with a set of real intervals. Then, we construct the graph, say G, of the membership function $y = m(x)$. There is a commonly used in fuzzy logic approach to measure performance with the coordinates $(x_c, y_c)$ of the *centre of gravity (centoid)*, say $F_c$, of the graph G, which we can calculate using the following well-known from Mechanics formulas:

$$x_c = \frac{\iint_F x\,dx\,dy}{\iint_F dx\,dy}, \quad y_c = \frac{\iint_F y\,dx\,dy}{\iint_F dx\,dy} \quad (1)$$

In our case we characterize a pair's performance as unsatisfactory (F), if $x \in [0, 1)$, as mediocre (D), if $x \in [1, 2)$, as good (C), if $x \in [2, 3)$, as very good (B), if $x \in [3, 4)$ and as excellent (A), if $x \in [4, 5]$ respectively. In other words, if $x \in [0, 1)$, then $y_1 = m(x) = m(F) = \frac{n_F}{n}$, if $x \in [1, 2)$, then $y_2 = m(x) = m(D) = \frac{n_D}{n}$, etc.

Therefore in our case the graph G of the membership function attached to S is the bar graph of Figure 1 consisting of five rectangles, say $G_i$, i=1,2,3, 4, 5, whose sides lying on the X axis have length 1. In this case $\iint_F dx\,dy$ is the area of G which is equal to $\sum_{i=1}^{5} y_i = \frac{n_F + n_D + n_C + n_B + n_A}{n} = 1$

Also $\iint_F x\,dx\,dy = \sum_{i=1}^{5} \iint_{F_i} x\,dx\,dy = \sum_{i=1}^{5} \int_0^{y_i} dy \int_{i-1}^{i} x\,dx$
$= \sum_{i=1}^{5} y_i \int_{i-1}^{i} x\,dx = \frac{1}{2} \sum_{i=1}^{5} (2i-1) y_i$, and

$$\iint_F y\,dxdy = \sum_{i=1}^{5}\iint_{F_i} y\,dxdy = \sum_{i=1}^{5}\int_0^{y_i} y\,dy \int_{i-1}^{i} dx =$$

$$\sum_{i=1}^{n}\int_0^{y_i} y\,dy = \frac{1}{2}\sum_{i=1}^{n} y_i^2 \qquad (1a)$$

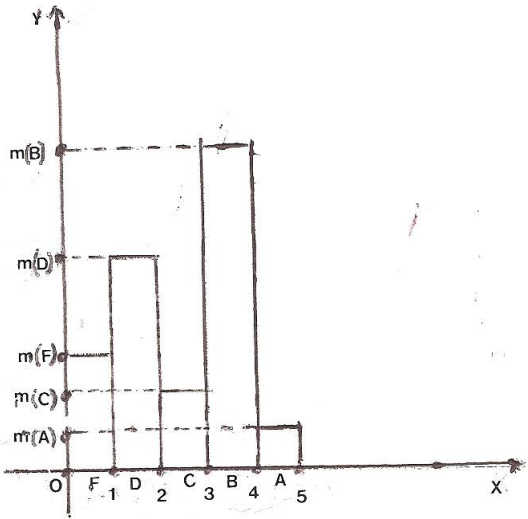

**Figure 1**: Bar graphical data representation

Therefore, using the relations (1a), formulas (1) are transformed into the following form:

$$x_c = \frac{1}{2}(y_1 + 3y_2 + 5y_3 + 7y_4 + 9y_5),$$

$$y_c = \frac{1}{2}(y_1^2 + y_2^2 + y_3^2 + y_4^2 + y_5^2) \qquad (2)$$

But, $0 \leq (y_1-y_2)^2 = y_1^2 + y_2^2 - 2y_1y_2$, therefore $y_1^2 + y_2^2 \geq 2y_1y_2$, with the equality holding if, and only if, $y_1=y_2$. In the same way one finds that $y_1^2 + y_3^2 \geq 2y_1y_3$, and so on. Hence it is easy to check that $(y_1+y_2+y_3+y_4+y_5)^2 \leq 5(y_1^2+y_2^2+y_3^2+y_4^2+y_5^2)$, with the equality holding if, and only if, $y_1=y_2=y_3=y_4=y_5$. But $y_1+y_2+y_3+y_4+y_5 =1$, therefore $1 \leq 5(y_1^2+y_2^2+y_3^2+y_4^2+y_5^2)$ (3), with the equality holding if, and only if, $y_1=y_2=y_3=y_4=y_5=\frac{1}{5}$.

Then the first of formulas (2) gives that $x_c = \frac{5}{2}$. Further, combining the inequality (3) with the second of formulas (2), one finds that $1 \leq 10y_c$, or $y_c \geq \frac{1}{10}$. Therefore the unique minimum for $y_c$ corresponds to the centre of gravity $F_m(\frac{5}{2}, \frac{1}{10})$.

The ideal case is when $y_1=y_2=y_3=y_4=0$ and $y_5=1$. Then from formulas (2) we get that $x_c = \frac{9}{2}$ and $y_c = \frac{1}{2}$. Therefore the centre of gravity in this case is the point $F_i(\frac{9}{2}, \frac{1}{2})$.

On the other hand, in the worst case $y_1=1$ and $y_2=y_3=y_4=y_5=0$. Then by formulas (2), we find that the centre of gravity is the point $F_w(\frac{1}{2}, \frac{1}{2})$.

Therefore the "area" where the centre of gravity $F_c$ lies is represented by the triangle $F_w F_m F_i$ of Figure 2.

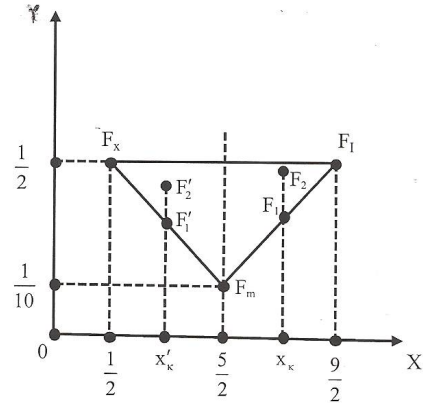

**Figure 2**: Graphical representation of the "area" of the centre of gravity

Then from elementary geometric considerations it follows that the greater is the value of $x_c$ the better is the corresponding group's performance. Also, for two groups with the same $x_c \geq 2,5$, the group having the centre of gravity which is situated closer to $F_i$ is the group with the higher $y_c$; and for two groups with the same $x_c < 2.5$ the group having the centre of gravity which is situated farther to $F_w$ is the group with the lower $y_c$. Based on the above considerations it is logical to formulate our criterion for comparing the groups' performances in the following form:

- *Among two or more groups the group with the higher $x_c$ performs better.*
- *If two or more groups have the same $x_c \geq 2.5$, then the group with the higher $y_c$ performs better.*
- *If two or more groups have the same $x_c < 2.5$, then the group with the lower $y_c$ performs better.*

## 3. Real applications

The *Hellenic Bridge Federation (HBF)* organizes, on a regular basis, *simultaneous* bridge tournaments (pair events) with pre-dealt boards, played by the local clubs in several cities of Greece. Each of these tournaments consists of six in total events, played in a particular day of the week (e.g. Wednesday), for six successive weeks. In each of these events there is a local scoring table (match points) for each participating club, as well as a central scoring table, based on the local results of all participating clubs, which are compared to each other. At the end of the tournament it is also formed a total scoring table in each club, for each player individually. In this table each player's score equals to the mean of the scores obtained by him/her in the five of the six in total events of the tournament. If a player has participated in all the events, then his/her worst score is dropped out. On the contrary, if he/she has participated in less than five events, his/her name is not included in this table and no possible extra bonuses are awarded to him/her.

In this section and in order to illustrate the importance of our results obtained in the previous section, we shall present two real applications connected to the above simultaneous tournaments.

The first application concerns the third event of such a simultaneous tournament played on Wednesday, March 12, 2014, in which participated 17 in total clubs from several cities of Greece (see results in [5]). Among those clubs were included the two bridge clubs, lets call them $C_1$ and $C_2$ respectively, of the city of Patras. Nine in total pairs from club $C_1$ played in this event obtaining the following scores in the central scoring table: 62.67%, 57.94%, 56.04%, 55.28%, 50.43%, 46%, 44.75%, 39.91% and 36.16%. Eight in total pairs from club $C_2$ played also in the same event obtaining the following scores: 63.14%, 57.64%, 56.86%, 50.17%, 50.13%, 43.28%, 42.11% and 36.63%. The above scores give an average percentage 49.909% for the first and 49.995% for the second club. This means that the second club demonstrated a slightly better mean performance than the first one, but the difference was marginal; only 0.086%.

The above results are summarized in Table 1.

**Table 1:** Results of the two bridge clubs of Patras

First club ($C_1$)

| % Scale | Performance | Amount of pairs | m(x) |
|---|---|---|---|
| >65% | A | 0 | 0 |
| 55-65% | B | 4 | 4/9 |
| 48-55% | C | 1 | 1/9 |
| 40-48% | D | 2 | 2/9 |
| <40% | F | 2 | 2/9 |
| Total | | 9 | |

Second club ($C_2$)

| % Scale | Performance | Amount of pairs | M(x) |
|---|---|---|---|
| >65% | A | 0 | 0 |
| 55-65% | B | 3 | 3/8 |
| 48-55% | C | 2 | 2/8 |
| 40-48% | D | 2 | 2/8 |
| <40% | F | 1 | 1/8 |
| Total | | 8 | |

Then, using the first of formulas (2) of the previous section one finds that $x_c = \frac{1}{2}(\frac{2}{9} + 3.\frac{2}{9} + 5.\frac{1}{9} + 7.\frac{4}{9}) = \frac{41}{18} \simeq 2.278$ for the first club, and $x_c = \frac{1}{2}(\frac{1}{8} + 3.\frac{2}{8} + 5.\frac{2}{8} + 7.\frac{3}{8}) = \frac{38}{16} = 2.375$ for the second club. Therefore, according to our criterion (first case) stated in the previous section, the second club demonstrated a better weighted performance than the first one, but the difference is small again; just 0.097 units.

The second application is related to the total scoring table of the players of club $C_1$, who participated in at least five of the six in total events of another simultaneous tournament organized by the HBF, which ended on February 19, 2014 (see results in [5]). Nine men and five women players are included in this table, who obtained the following scores. Men: 57.22%, 54.77%, 54.77%, 54.35%, 54.08%, 50.82%, 50.82%, 49.61%, 47.82%. Women: 59.48%, 54.08%, 53.45%, 53.45%, 47.39%. The above results give a mean percentage of approximately 52.696% for the men and 53.57% for the women players. Therefore the women demonstrated a slightly better mean performance than the men players, their difference being 0.874%.

The above results are summarized in Table 2.

**Table 2:** Total scoring of the men and women players

Men

| % Scale | Performance | Amount of players | m(x) |
|---|---|---|---|
| >65% | A | 0 | 0 |
| 55-65% | B | 1 | 1/9 |
| 48-55% | C | 7 | 7/9 |
| 40-48% | D | 1 | 1/9 |
| <40% | F | 0 | 0 |
| Total | | 9 | |

Women

| % Scale | Performance | Amount of players | m(x) |
|---|---|---|---|
| >65% | A | 0 | 0 |
| 55-65% | B | 1 | 1/5 |
| 48-55% | C | 3 | 3/5 |
| 40-48% | D | 1 | 1/5 |
| <40% | F | 0 | 0 |
| Total | | 5 | |

Thus, according to the first of formulas (2) of the previous section, we find that $x_c = \frac{1}{2}(3.\frac{1}{9} + 5.\frac{7}{9} + 7.\frac{1}{9}) = \frac{45}{18} = 2.5$ for the men players, and $x_c = \frac{1}{2}(3.\frac{1}{5} + 5.\frac{3}{5} + 7.\frac{1}{5}) = \frac{25}{10} = 2.5$ for the women players. Further, the second of formulas (2) gives $y_c = \frac{1}{2}[(\frac{1}{9})^2 + (\frac{7}{9})^2 + (\frac{1}{9})^2] = \frac{51}{162} \simeq 0.315$ for the men and $y_c = \frac{1}{2}[(\frac{1}{5})^2 + (\frac{3}{5})^2 + (\frac{1}{5})^2] = \frac{11}{50} = 0.22$ for the women players. Thus, according to our criterion (second case) and in contrast to the mean performance, the men demonstrated a higher weighted performance than the women players.

## 4. Conclusions and discussion

In the present paper we developed a new method for assessing the total performance of certain groups of pairs or teams or of bridge players individually, appearing to have a special interest. In developing the above method we represented each of the groups of players' under assessment as a fuzzy subset of a set $U$ of linguistic labels characterizing the bridge players' performance and we used the centroid defuzzification technique in converting the fuzzy data collected from the game to a crisp number. According to the above assessment method the higher is an element's performance the more its "contribution" to the corresponding set's total performance (weighted performance). Thus, in contrast to the mean of the scores of all set's elements, which is connected to the mean group's performance, our method is connected somehow to the group's *quality performance*. As a result, when the above two different assessment methods are used in comparing the performance of two or more groups of pairs/teams of bridge players, the results obtained may differ to each other in certain cases, where there are marginal differences in the groups' performance.

Two real applications were also presented, related to simultaneous tournaments (pair events) organized by the HBF. In the first of these applications we compared the total performance of the two bridge clubs of the city of Patras in a particular event of a recent such tournament, while in the second one we compared the performance of the men and women players of one of the above clubs, based on their total scoring in the six events of another simultaneous tournament.

In general, our method is suitable to be applied in parallel with the official bridge scoring methods (match points or IMPs) for statistical and other obvious reasons.

Our future plans for further research on the subject aim at applying our new assessment method in more real situations, including also bridge games (pairs or teams) played with IMPs, in order to get statistically safer and more solid conclusions about its applicability and usefulness. In a wider basis, since our method is actually a general assessment method, it could be extended to cover other sectors of the human activity as well, apart from the students' (e.g. see [10], [12-13], [16-18], etc) and the bridge players' assessment (in this paper), where we have already applied it.

## Acknowledgement

The author wishes to thank Dr. Athanasios Tsevis, Chemical Engineer, Hellenic Open University, Patras, Greece, for introducing him to the principles and techniques of the game of bridge.